\documentclass[]{spie}  

\usepackage{amsmath,amsfonts,amssymb}
\usepackage{subcaption}
\usepackage{graphicx}
\usepackage{cite} 
\usepackage{epsfig}
\usepackage{nccmath}
\usepackage{xcolor,colortbl}
\usepackage{tabularx}
\usepackage{relsize}
\usepackage{pifont}
\newcommand{\xmark}{\ding{55}}
\usepackage{booktabs} 
\usepackage{multirow}
\usepackage{multicol}
\usepackage{adjustbox}
\usepackage{float}
\usepackage{makecell}
\usepackage{tabu}
\usepackage[colorlinks=true, allcolors=blue]{hyperref}
\usepackage[capitalize]{cleveref}
\usepackage{algorithm}

\title{How Close Are We? Limitations and Progress of AI Models in Banff Lesion Scoring}

\author[a]{Yanfan Zhu}
\author[a]{Juming Xiong}
\author[b]{Ruining Deng}
\author[c]{Yu Wang}
\author[d]{Yaohong Wang}
\author[c]{Shilin Zhao}
\author[c]{Mengmeng Yin}
\author[e]{Yuqing Liu}
\author[c]{Haichun Yang}
\author[a]{Yuankai Huo}

\affil[a]{Vanderbilt University, Nashville TN 37235, USA}
\affil[b]{Weill Cornell Medicine, New York, NY 10021, USA}
\affil[c]{Vanderbilt University Medical Center, Nashville TN 37232, USA}
\affil[d]{UT MD Anderson Cancer Center,TX 77030, USA}
\affil[e]{Tongji University School of Medicine, Shanghai, 200092, China}

\authorinfo{Further author information: (Send correspondence to Yuankai Huo)\\
E-mail: yuankai.huo@vanderbilt.edu}

\begin{document}
\maketitle

\begin{abstract}
The Banff Classification provides the global standard for evaluating renal transplant biopsies, yet its semi-quantitative nature, complex criteria, and inter-observer variability present significant challenges for computational replication. In this study, we explore the feasibility of approximating Banff lesion scores using existing deep learning models through a modular, rule-based framework. We decompose each Banff indicator—such as glomerulitis (\textit{g}), peritubular capillaritis (\textit{ptc}), and intimal arteritis (\textit{v})—into its constituent structural and inflammatory components, and assess whether current segmentation and detection tools can support their computation. Model outputs are mapped to Banff scores using heuristic rules aligned with expert guidelines, and evaluated against expert-annotated ground truths. Our findings highlight both partial successes and critical failure modes, including structural omission, hallucination, and detection ambiguity. Even when final scores match expert annotations, inconsistencies in intermediate representations often undermine interpretability. These results reveal the limitations of current AI pipelines in replicating computational expert-level grading, and emphasize the importance of modular evaluation and computational Banff grading standard in guiding future model development for transplant pathology.

\end{abstract}


\section{Introduction}
The \textit{Banff Classification of Allograft Pathology} has served as the international standard for evaluating renal transplant biopsies since 1991. With biennial updates and expert summits, it plays a central role in unifying diagnostic protocols across transplant centers.However, the schema’s evolving nature has led to fragmented documentation over time, making consistent interpretation difficult in both clinical and research settings. Moreover, due to the semi-quantitative nature of Banff scoring and lack of standardized visual references, significant inter-observer variability exists even among expert pathologists. To address this, the Banff Foundation recently launched a centralized digital repository consolidating lesion definitions, scoring criteria, and diagnostic categories~\cite{banffrepo2023}. This complexity and constant evolution present major barriers to computational reproducibility and standardization.

Although Banff scoring aims to be systematic, its real-world application often varies across institutions and pathologists, reflecting inherent subjectivity in lesion interpretation. The rise of whole slide imaging (WSI) and AI-powered pathology has prompted increasing interest in automating Banff lesion scoring. Yet, challenges remain, notably the lack of large-scale datasets annotated according to Banff definitions and the histological complexity of transplant pathology. Efforts like the DIAGGRAFT challenge by the Banff Digital Pathology Working Group (DPWG) have taken steps toward developing inflammation detection models and integrating them with segmentation tools~\cite{schinstock2023banffdpwg}.

Notably, the Banff 2022 Working Group conducted pilot studies using Mask R-CNN~\cite{he2017maskrcnn} and U-Net~\cite{ronneberger2015unet} to segment inflammatory lesions on PAS-stained slides~\cite{yi2022deep}. These early explorations confirmed the technical potential of AI for Banff tasks, especially in glomerular and inflammation-rich regions, while highlighting the need for more structured integration of lesion definitions and computational targets.

Building upon these foundations, our study does not seek to develop a complete Banff scoring solution, but rather to explore the practical feasibility of approximating Banff lesion scores using existing AI models. We construct a modular pipeline that integrates structural segmentation and inflammatory cell detection, aiming to assess which components of Banff scoring can be supported by current tools. While our focus is on feasibility rather than accuracy, this framework provides a foundation for systematically identifying both capabilities and limitations in current AI-driven pathology workflows. Recent studies have suggested that certain quantitative image features—such as inflammatory cell density and spatial distribution—may correlate with Banff scores, particularly for inflammation-associated indicators like \textit{i} (interstitial inflammation) and \textit{t} (tubulitis). For example, Yi et al.~\cite{yi2024large} applied automated inflammation quantification to renal biopsies and reported promising agreement with expert annotations. However, such findings primarily demonstrate correlation rather than end-to-end scoring validity, and their generalizability across different lesion types and tissue contexts remains uncertain.

\begin{figure}[t!]
\centering
\includegraphics[width=1\textwidth]{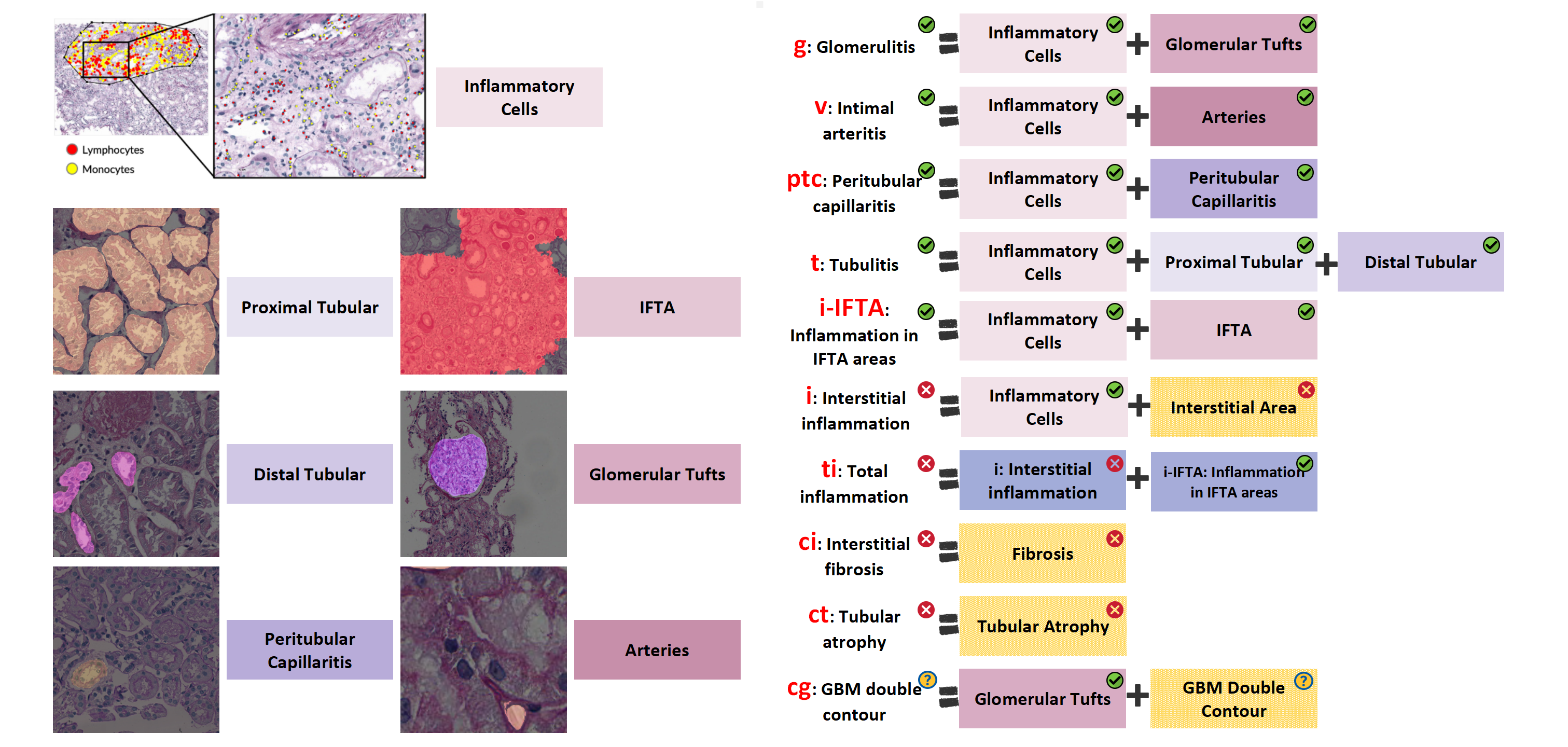}
\caption{
Component-level decomposition of Banff lesion scores and model support status.
Each Banff indicator (e.g., \textit{g}, \textit{v}, \textit{ptc}) is represented as a combination of required cellular and structural features. The left panel shows visual examples of representative tissue components segmented by existing models. The right panel maps each Banff score to its constituent elements and indicates whether current models in our pipeline can detect them: \checkmark{} indicates sufficient support, \xmark{} indicates missing capability, and ? indicates partial or uncertain feasibility. This mapping reveals which Banff scores are currently tractable using existing tools and highlights unresolved challenges, especially for indicators like \textit{i}, \textit{ci}, \textit{ct}, and \textit{cg}.
}
\label{fig:overview}
\end{figure}

To assess the computational feasibility of automating Banff lesion assessment, we systematically deconstructed each Banff indicator into its requisite cellular and structural components based on formal definitions from the Banff Foundation. As illustrated in Figure~\ref{fig:overview}, many lesion types—such as glomerulitis (\textit{g}), intimal arteritis (\textit{v}), and peritubular capillaritis (\textit{ptc})—require both inflammatory cell detection and the segmentation of associated microanatomical regions (e.g., glomerular tufts, arteries). Other indicators, such as \textit{t} (tubulitis) and \textit{i-IFTA} (inflammation in fibrotic areas), demand region-aware quantification of inflammation within specific tissue compartments.

Structural lesions like \textit{ci} (interstitial fibrosis) and \textit{ct} (tubular atrophy) require more fine-grained parsing of fibrotic and atrophic zones within the broader IFTA label. In the case of \textit{cg} (glomerular basement membrane double contour), although specialized models such as GBMSeg~\cite{liu2024featurepromptinggbmsegoneshotreference} have been proposed, they operate at the patch level and do not yet support whole-slide scoring.

We evaluated each component for model feasibility using publicly available detection and segmentation tools. Features marked with \checkmark{} in Figure~\ref{fig:overview} are currently supported; those marked with \xmark{} lack model coverage due to insufficient data, lack of task-specific training, or poorly defined region boundaries. The \textbf{?} symbol highlights components that may be partially detectable, but where model performance, robustness, or definition alignment remains uncertain—such as GBM contours in \textit{cg} scoring.

Rather than presenting a comprehensive solution, this decomposition serves to map the feasibility boundaries of current AI-based workflows. It highlights which Banff indicators are currently tractable and which remain unresolved, offering a roadmap for prioritizing future model development and annotation efforts.

While no existing AI pipeline fully aligns with the Banff scoring system, several recent models offer partial support for key lesion components. For example, the Monkey Challenge model~\cite{monkeychallenge2023} detects and classifies inflammatory cells relevant to \textit{i}, \textit{t}, and \textit{ptc} scoring. Omni-Seg~\cite{deng2023omnisegscaleawaredynamicnetwork} enables structural segmentation of glomeruli, tubules, and vessels—essential for evaluating \textit{g}, \textit{ptc}, and \textit{v}. Glo-in-One V2~\cite{yu2024gloinonev2holisticidentificationglomerular} provides instance-level glomerular detection supporting \textit{g} and \textit{cg}, while IRS~\cite{deng2025irsincrementalrelationshipguidedsegmentation} incorporates immune-aware features potentially useful for vascular lesion modeling.

Despite these individual capabilities, none of these tools can independently support full Banff scoring. To bridge this gap, we adopt a component-level integration strategy—not to construct a unified scoring system, but to evaluate how existing outputs can approximate individual Banff indicators. By decomposing each score into its constituent structural and inflammatory elements, we construct a feasibility map that reveals which scores are currently tractable and where significant limitations persist.

\vspace{0.5em}
\noindent
Our contributions are summarized as follows:

\begin{itemize}
  \item We deconstruct each Banff lesion score into its underlying structural and inflammatory components, enabling a fine-grained evaluation of which elements are currently computable using existing AI tools. Rather than focusing on final score prediction, we assess the detectability of intermediate features (e.g., glomeruli, arteries, inflammatory cells) and their relevance to specific Banff indicators (Figure~\ref{fig:overview}).

  \item We implement a modular evaluation framework that combines region-of-interest (ROI) detection, tissue structure segmentation, and inflammatory cell prediction. This setup allows us to approximate selected Banff scores—specifically \textit{g}, \textit{v}, and \textit{ptc}—and compare these predictions to expert annotations curated by the Banff Foundation, thereby identifying both strengths and limitations of current model outputs in practice.
\end{itemize}

\section{Methods}

Our goal is not to build a unified Banff scoring system, but rather to evaluate whether existing AI models can provide sufficient anatomical and cellular information to support Banff lesion assessment. We assemble a modular evaluation pipeline that integrates off-the-shelf or pretrained models for three key tasks: (1) tissue section detection, (2) structural segmentation, and (3) inflammatory cell detection. These modules provide the basic inputs required by Banff scoring rules.

We then define indicator-specific rules that map model outputs to lesion scores, allowing us to evaluate model feasibility on a per-indicator basis. Importantly, our approach decouples model development from lesion scoring, enabling a clear analysis of where existing tools succeed or fail in reproducing Banff criteria.t.

\subsection{Modular Composition of Existing AI Tools}

To assess whether current AI methods can provide sufficient support for Banff lesion assessment, we constructed a modular pipeline by integrating several pretrained models targeting distinct tissue and cellular components. While not intended as a finalized scoring system, this configuration enables component-level evaluation of model capabilities in relation to Banff indicators.

We directly processed pre-cropped biopsy fragments obtained from the WSIs. These fragments correspond to representative tissue sections either manually selected or provided as pre-defined ROIs, serving as the entry point for subsequent segmentation and detection modules. For structural segmentation, we employed Omni-Seg~\cite{deng2023omnisegscaleawaredynamicnetwork}, a general-purpose semantic segmentation model originally designed for multi-organ analysis. We adapted its inference pipeline to process full-resolution kidney biopsy sections in batch mode, producing pixel-wise maps of anatomical structures including glomeruli, proximal and distal tubules, arteries, and interstitial regions. Inflammatory cell detection was performed using a two-stage model stack adapted from the Monkey Challenge~\cite{monkeychallenge2023}. A YOLOv11 detector was first applied to image patches ($1024 \times 1024$ pixels) to localize candidate lymphocytes and monocytes. These predictions were refined via a second-stage classification step, using EfficientNet-B0~\cite{tan2019efficientnet} classifiers trained to filter false positives. Verified cell coordinates were stored in structured JSON format for downstream mapping.

Rather than retraining or fine-tuning these models, we used their out-of-the-box outputs to test whether they could produce the anatomical and cellular elements required for Banff score computation. This setup provides a practical and reproducible framework for assessing the current limitations and feasibility boundaries of AI-assisted Banff lesion quantification. In the next stage, we applied rule-based mappings to evaluate whether these outputs, when combined, suffice to compute selected Banff lesion indicators in alignment with expert guidelines.

\subsection{Lesion Mapping and Banff Indicator Estimation}

\begin{figure}[ht!]
    \centering
    \begin{subfigure}{\textwidth}
        \centering
        \includegraphics[width=\textwidth]{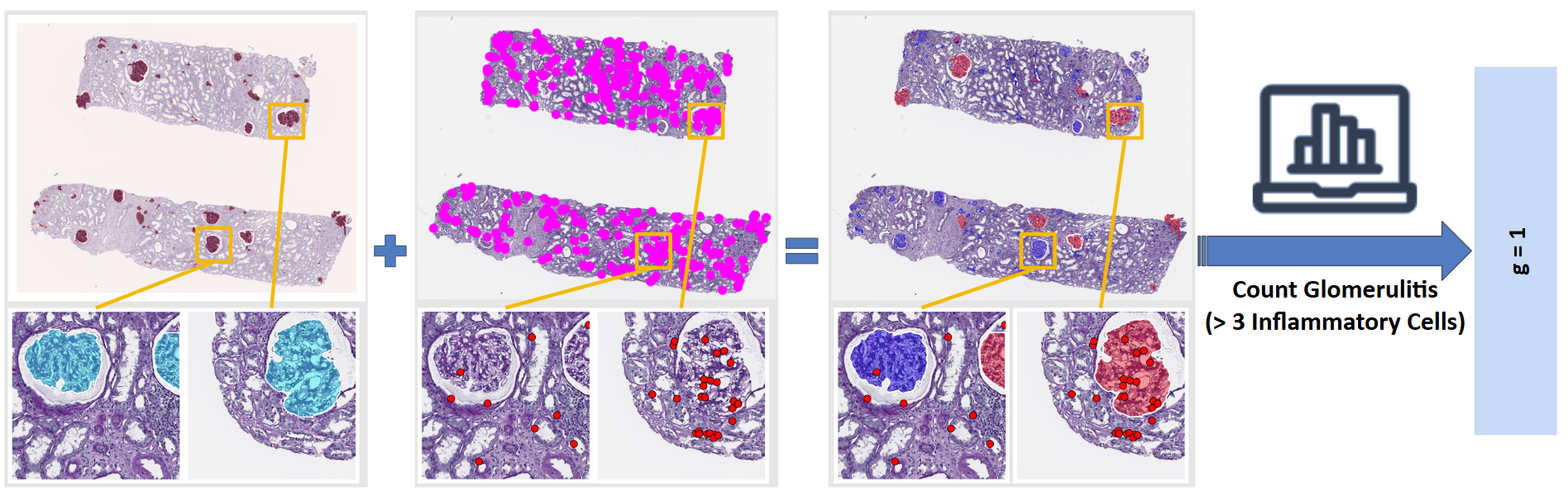}
        \caption{Glomerulitis ($g$): Count the number of glomeruli containing more than three inflammatory cells. The final $g$ score is computed according to Eq.~\ref{eq:g-score}.}
        \label{fig:banff_g}
    \end{subfigure}

    \vspace{1em}

    \begin{subfigure}{\textwidth}
        \centering
        \includegraphics[width=\textwidth]{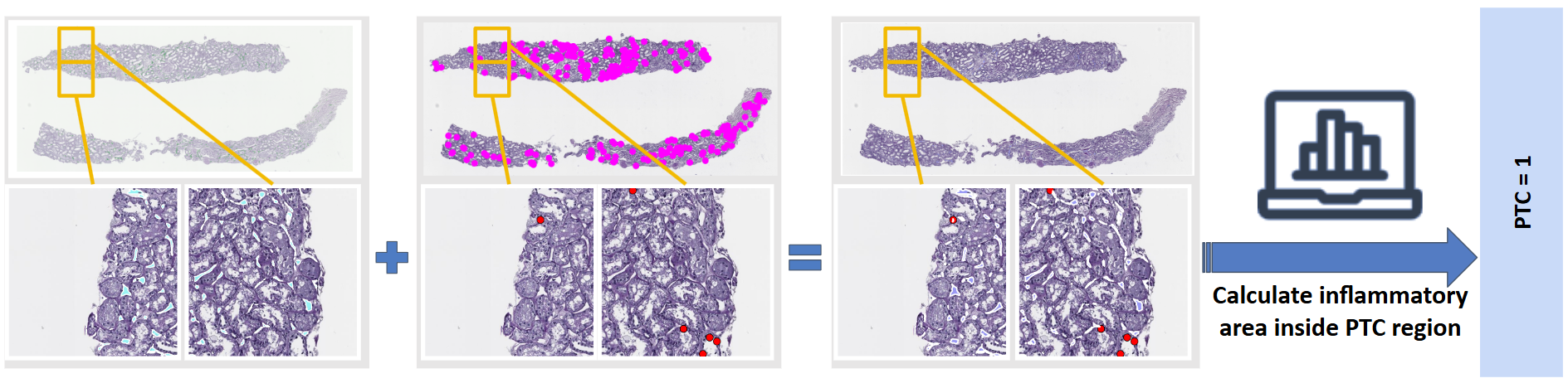}
        \caption{Peritubular capillaritis ($ptc$): Estimate the number of inflammatory cells within PTC regions and assign scores based on the maximum cell count, as defined in Eq.~\ref{eq:ptc-score}.}
        \label{fig:banff_ptc}
    \end{subfigure}

    \vspace{1em}

    \begin{subfigure}{\textwidth}
        \centering
        \includegraphics[width=\textwidth]{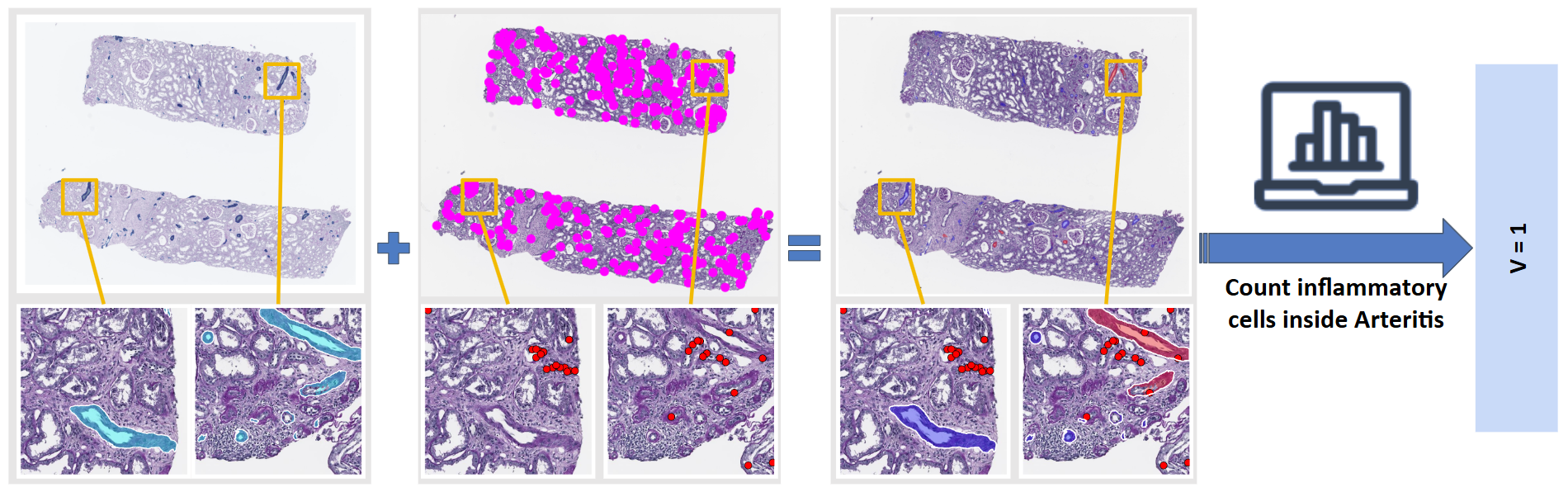}
        \caption{Intimal arteritis ($v$): Detect inflammatory cells in arterial regions and determine the score based on their maximal count per artery, following Eq.~\ref{eq:v-score}.}
        \label{fig:banff_v}
    \end{subfigure}

    \caption{
        Visual workflow for Banff lesion score computation. Each row illustrates the process of combining structural segmentation and inflammatory cell detection to derive the Banff score for (a) glomerulitis, (b) peritubular capillaritis, and (c) intimal arteritis. The final scores are assigned based on the corresponding rules and thresholds defined in Eqs.~\ref{eq:g-score},~\ref{eq:ptc-score}, and~\ref{eq:v-score}.
    }
    \label{fig:banff-score-computation}
\end{figure}

Following structural segmentation and inflammatory cell detection, we compute three Banff lesion scores—glomerulitis ($g$), peritubular capillaritis ($ptc$), and intimal arteritis ($v$)—by assigning verified inflammatory cells to instance-segmented anatomical regions (e.g., glomerular tufts, peritubular capillaries, arteries). Each lesion score is computed through per-instance aggregation of inflammatory burden, followed by heuristic thresholds derived from Banff definitions. These rule sets allow us to test whether current model outputs suffice for lesion quantification in a compartment-specific manner. 

Figure~\ref{fig:banff-score-computation} visually illustrates the lesion scoring process, where each indicator is derived by combining instance-level predictions from structure and cell models. Final scores are assigned using the rules defined in Eqs.~\ref{eq:g-score},~\ref{eq:ptc-score}, and~\ref{eq:v-score}, respectively. This instance-based approach enables compartment-specific lesion estimation and aligns with pathologist-level lesion grading.

\paragraph{Glomerulitis ($g$).}
Let $\{ R_1, R_2, \dots, R_N \}$ denote the set of segmented glomerular tufts. For each tuft $R_k$, define the count of inflammatory cells:
\[
n_k = \sum_{p_i \in \mathcal{P}} \mathbb{I}[p_i \in R_k]
\]
Then define the indicator variable:
\[
\delta_k = 
\begin{cases}
1, & \text{if } n_k > 3 \\
0, & \text{otherwise}
\end{cases}
\]
where $\mathcal{P}$ denotes the set of all inflammatory cell detections. The proportion of inflamed glomeruli is:
\[
\rho_g = \frac{1}{N} \sum_{k=1}^{N} \delta_k
\]
The Banff $g$ score is then defined as:
\begin{equation}
g = 
\begin{cases}
0, & \rho_g = 0 \\
1, & 0 < \rho_g < 0.25 \\
2, & 0.25 \leq \rho_g \leq 0.5 \\
3, & \rho_g > 0.5
\end{cases}
\label{eq:g-score}
\end{equation}

\paragraph{Peritubular Capillaritis ($ptc$).}
Let $\{ T_1, T_2, \dots, T_K \}$ be the set of peritubular capillary (PTC) instances. For each $T_k$, define the local inflammatory cell count:
\[
n_k = \sum_{p_i \in \mathcal{P}} \mathbb{I}[p_i \in T_k]
\]
where $\mathbb{I}[\cdot]$ is the indicator function. The maximum count across all PTCs is:
\[
\tilde{n} = \max\{ n_1, n_2, \dots, n_K \}
\]
The $ptc$ score is assigned as:
\begin{equation}
ptc = 
\begin{cases}
0, & \tilde{n} = 0 \\
1, & 1 \leq \tilde{n} \leq 4 \\
2, & 5 \leq \tilde{n} \leq 10 \\
3, & \tilde{n} > 10
\end{cases}
\label{eq:ptc-score}
\end{equation}

\paragraph{Intimal Arteritis ($v$).}
Similarly, for a set of arterial segments $\{ A_1, A_2, \dots, A_M \}$, we compute:
\[
m_j = \sum_{p_i \in \mathcal{P}} \mathbb{I}[p_i \in A_j], \quad \tilde{m} = \max\{ m_1, m_2, \dots, m_M \}
\]
The $v$ score is determined by:
\begin{equation}
v = 
\begin{cases}
0, & \tilde{m} = 0 \\
1, & 1 \leq \tilde{m} \leq 4 \\
2, & 5 \leq \tilde{m} \leq 10 \\
3, & \tilde{m} > 10
\end{cases}
\label{eq:v-score}
\end{equation}

This rule-based quantification scheme offers a transparent way to approximate Banff scoring by mapping inflammatory cells to anatomically defined compartments. While such structured mapping enables spatially grounded lesion estimates, the discrete thresholds used for final score assignment may oversimplify the nuanced, semi-quantitative nature of Banff criteria. Moreover, the accuracy of lesion scores depends heavily on the precision of both segmentation and detection modules. These factors collectively highlight the limitations of current AI pipelines in fully replicating expert-level lesion grading.

\section{Data and Experimental Setting}

To evaluate the proposed Banff lesion assessment workflow, we collected whole slide images (WSIs) from 20 renal transplant patients. All slides were stained with periodic acid--Schiff (PAS) and scanned at 40$\times$ magnification using a high-resolution digital scanner. Each WSI was annotated by renal pathologists to delineate regions of interest (ROIs), and lesion-level Banff scores were assigned to selected regions. It is important to note that Banff scores are inherently semi-quantitative and may vary between observers even for the same lesion, reflecting the interpretive nature of human pathology.

From each WSI, one representative PAS-stained section was selected based on visibility, completeness, and the presence of relevant lesion patterns, resulting in a total of 12 sections from 10 different patients for downstream analysis. These regions were either manually identified or selected based on slide metadata, without the use of automated tissue detection.

For each selected section, we applied the proposed multi-stage workflow to assess three Banff lesion categories: glomerulitis (\textit{g}), peritubular capillaritis (\textit{ptc}), and intimal arteritis (\textit{v}). The corresponding segmentation and detection outputs were mapped to lesion scores using pre-defined rules aligned with the Banff 2019 criteria. Ground-truth scores were extracted from expert-provided GeoJSON annotations encoded directly with lesion grades.

Since ground-truth annotations in our dataset are limited to ROIs manually selected by expert pathologists—rather than full-slide pixel-level labels—we focus our evaluation on lesion-level score comparison. Specifically, we compare model-predicted and expert-assigned categorical grades (0--3) using confusion matrices for each lesion type. These matrices provide an interpretable summary of model behavior, revealing agreement trends as well as systematic misclassifications.

Pixel-level metrics such as semantic or instance-level Dice scores are omitted from our primary analysis due to the partial nature of annotation coverage and our emphasis on end-to-end lesion scoring accuracy.

\section{Results}
\begin{figure}[!ht]
    \centering
    \begin{subfigure}{0.32\textwidth}
        \centering
        \includegraphics[width=\linewidth]{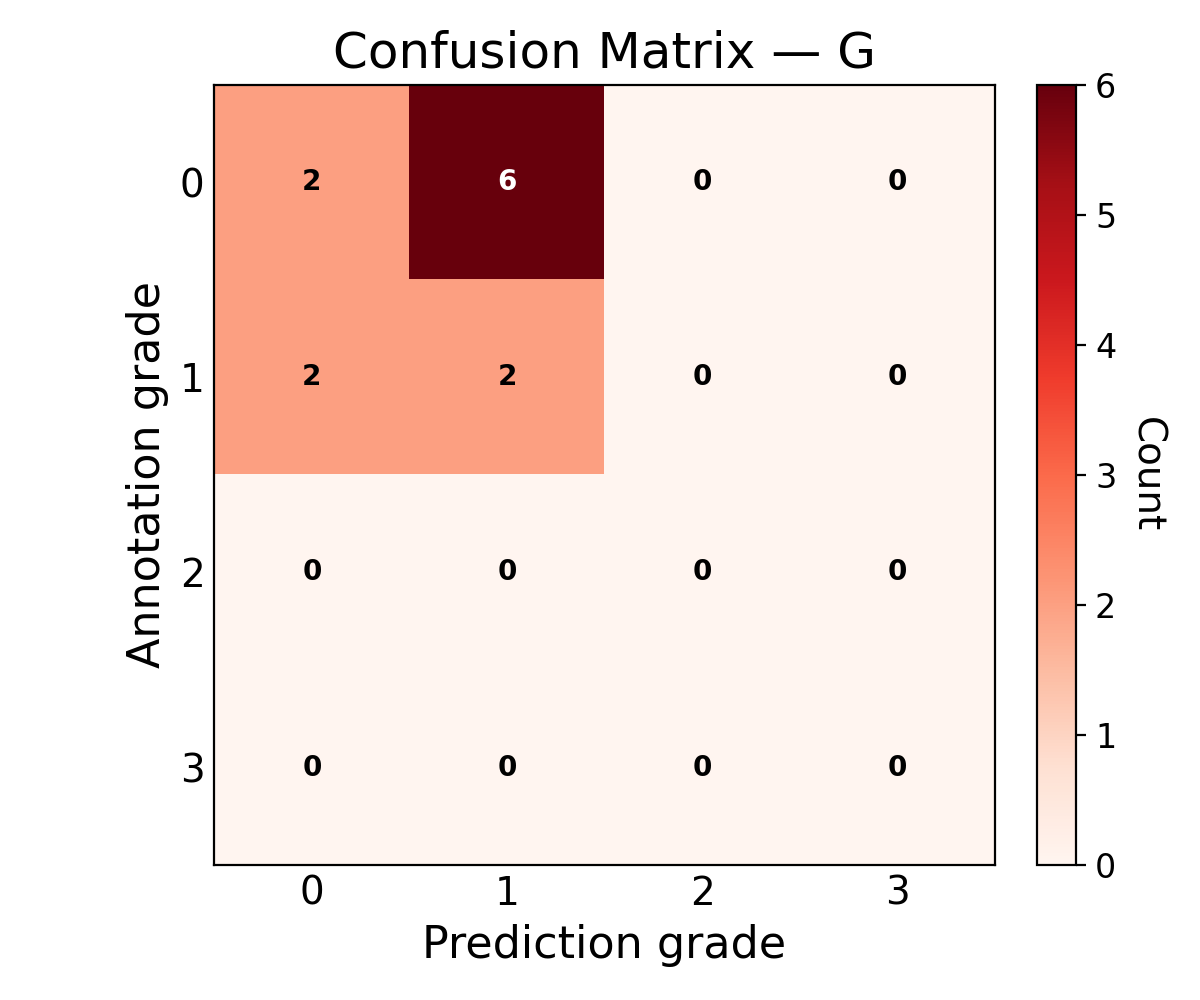}
        \caption{$g$ (glomerulitis)}
        \label{fig:cm-g}
    \end{subfigure}
    \hfill
    \begin{subfigure}{0.32\textwidth}
        \centering
        \includegraphics[width=\linewidth]{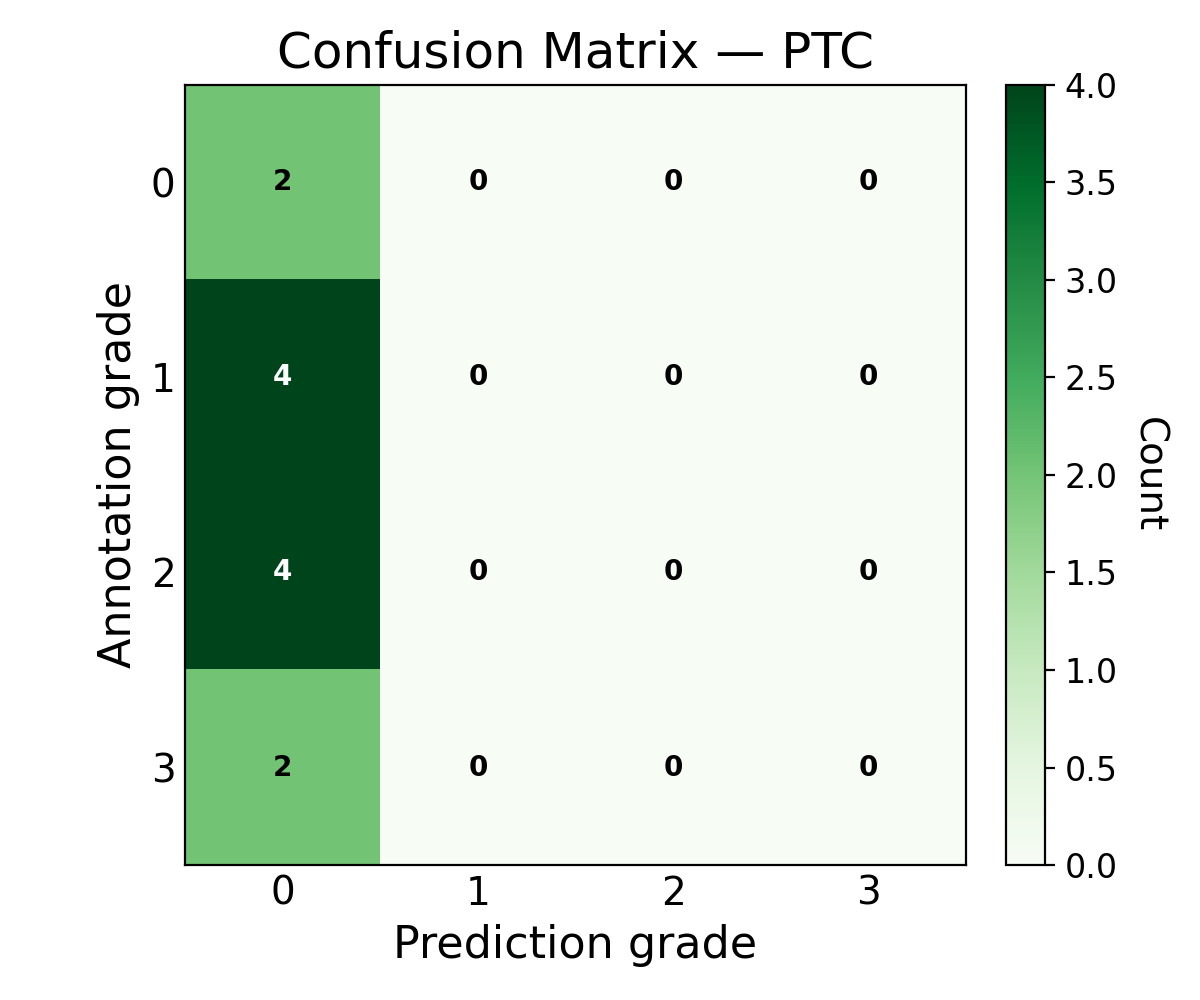}
        \caption{$ptc$ (peritubular capillaritis)}
        \label{fig:cm-ptc}
    \end{subfigure}
    \hfill
    \begin{subfigure}{0.32\textwidth}
        \centering
        \includegraphics[width=\linewidth]{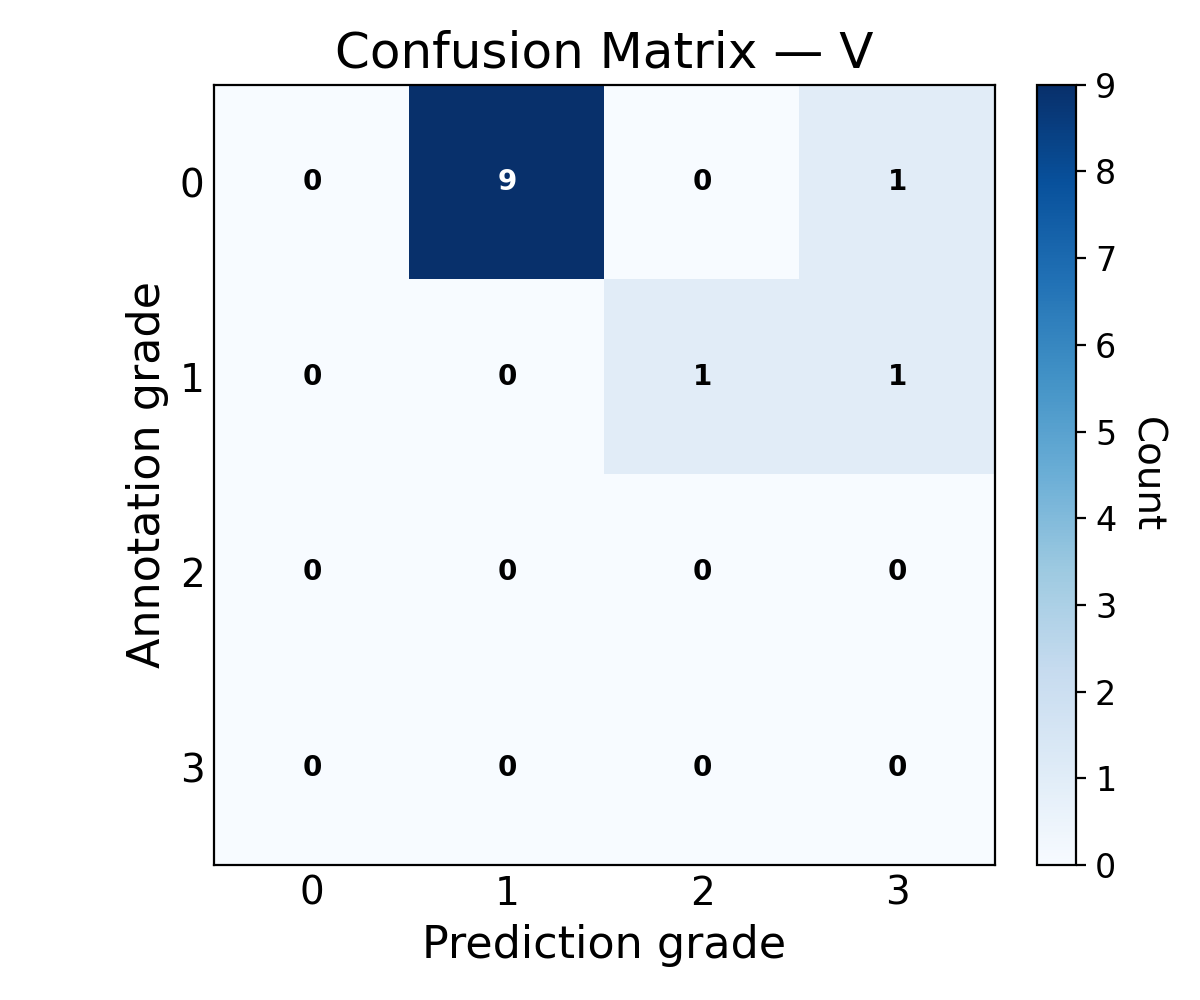}
        \caption{$v$ (intimal arteritis)}
        \label{fig:cm-v}
    \end{subfigure}
    \caption{
        Confusion matrices for model-predicted Banff scores versus ground truth across three lesion types: (a) glomerulitis ($g$), (b) peritubular capillaritis ($ptc$), and (c) intimal arteritis ($v$). Color intensity reflects prediction counts, with perfect agreement on the diagonal.
    }
    \label{fig:banff-cm}
\end{figure}

\begin{figure}[!ht]
    \centering
    \begin{subfigure}{0.48\textwidth}
        \centering
        \includegraphics[height=5cm]{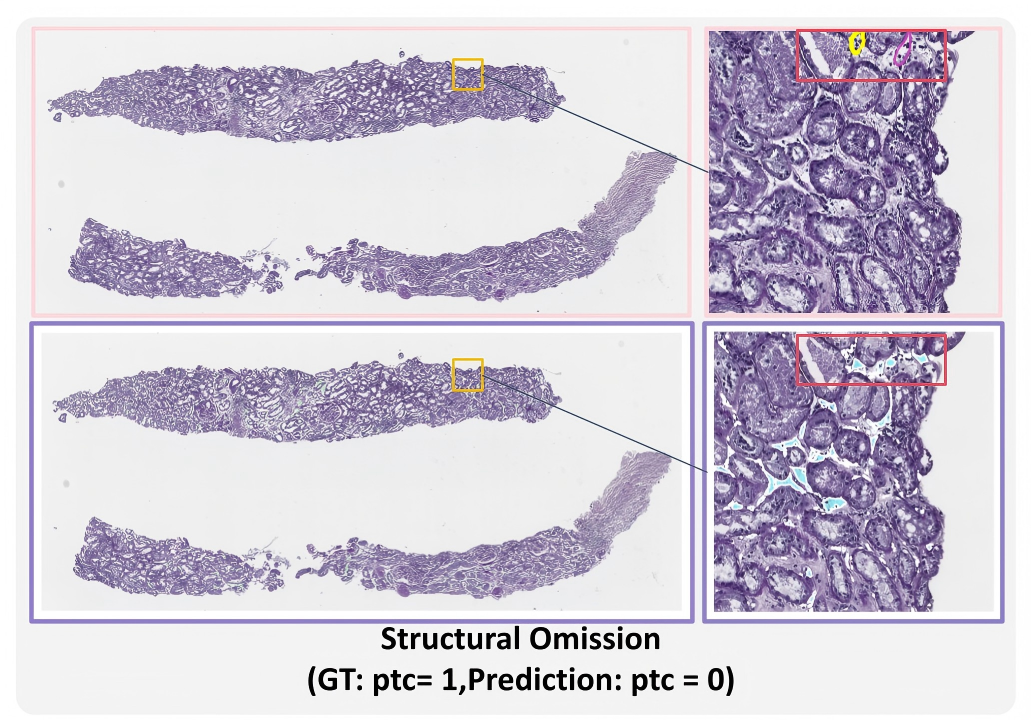}
        \caption{
            Structural omission: a peritubular capillary containing inflammatory cells is missed,
            leading to an underestimated \textit{ptc} score (GT: \textit{ptc}=1, Pred: \textit{ptc}=0).
        }
        \label{fig:structural-omission}
    \end{subfigure}
    \hfill
    \begin{subfigure}{0.48\textwidth}
        \centering
        \includegraphics[height=5cm]{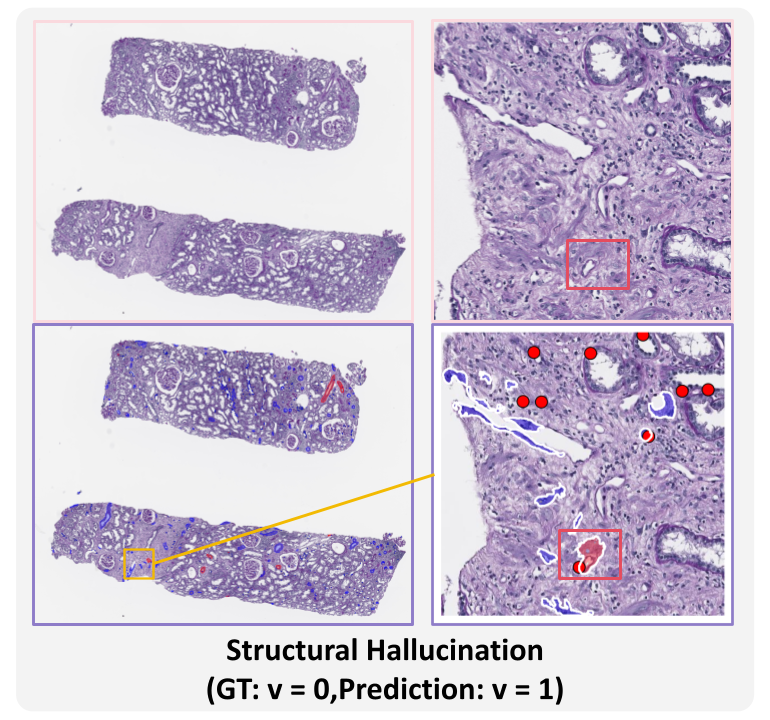}
        \caption{
            Structural hallucination: a non-existent arterial structure is falsely segmented and
            filled with inflammatory detections, producing an inflated \textit{v} score (GT: \textit{v}=0, Pred: \textit{v}=1).
        }
        \label{fig:structural-hallucination}
    \end{subfigure}
    \caption{
        Representative structural segmentation errors affecting Banff lesion scoring.
        (a) Omission errors cause false negatives in \textit{ptc}, while
        (b) hallucinations cause false positives in \textit{v}. 
        Both highlight how segmentation inaccuracies propagate to downstream Banff scores.
    }
    \label{fig:structural-failures}
\end{figure}

\begin{figure}[!ht]
    \centering
    \includegraphics[width=\textwidth]{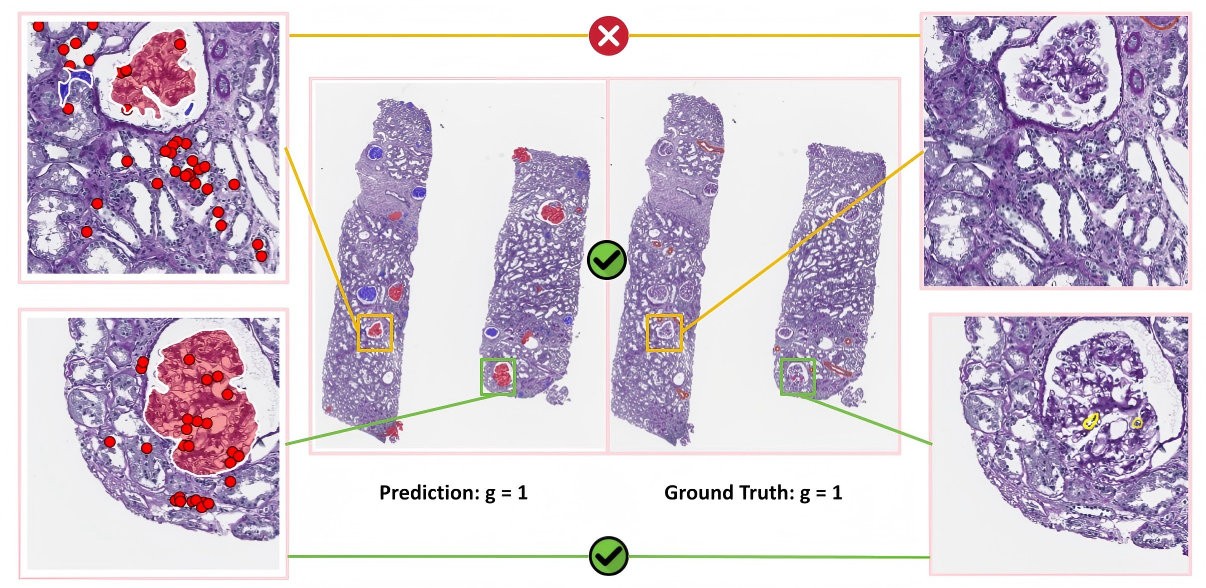}
    \caption{
        A representative case where the predicted glomerulitis score ($g=1$) matches the ground truth, yet the inflammatory cell detection results remain ambiguous. Left: model-predicted structural segmentation and inflammatory cells overlaid on the tissue section. Middle: zoomed-in glomerulus with annotated true positives (TP), false positives (FP), and false negatives (FN). Right: expert-annotated ground truth image. Although the overall score is correct, the segmentation and localization of inflammatory cells are not precisely aligned with the expert reference. This illustrates the inherent difficulty of reliably computing semi-quantitative Banff scores when intermediate representations—such as inflammatory cell maps—are themselves uncertain or ill-defined.
    }
    \label{fig:good-pred-but-uncertain}
\end{figure}

Figure~\ref{fig:banff-cm} shows confusion matrices for model-predicted versus expert-annotated Banff grades across $g$, $ptc$, and $v$. For $g$ (Figure~\ref{fig:cm-g}), predictions are concentrated in grades~0--1, with only partial agreement at lower grades and no detection of higher grades. For $ptc$ (Figure~\ref{fig:cm-ptc}), the model assigns nearly all cases to grade~0, regardless of annotation. For $v$ (Figure~\ref{fig:cm-v}), outputs are similarly skewed toward grades~0--1 with minimal identification of higher grades. These trends indicate systematic underestimation, likely due to upstream segmentation or detection errors, and underscore the difficulty of replicating Banff’s semi-quantitative grading using current AI models.

Figures~\ref{fig:structural-omission} and~\ref{fig:structural-hallucination} present representative structural segmentation errors that directly influence Banff score predictions. In Figure~\ref{fig:structural-omission}, a peritubular capillary is entirely missed by the segmentation model, producing a false negative for \textit{ptc}. In Figure~\ref{fig:structural-hallucination}, a non-existent arterial structure is falsely segmented and populated with inflammatory cell detections, leading to an inflated \textit{v} score.

Importantly, even when model predictions match the ground truth score, the underlying reasoning may remain uncertain. Figure~\ref{fig:good-pred-but-uncertain} shows a case where the predicted glomerulitis score is correct ($g=1$), yet the inflammatory cell detections within the glomerulus include both false positives and false negatives. Because final scores are derived from thresholded counts (e.g., $g=1$ if more than 3 cells are detected), small inconsistencies in detection near decision boundaries can lead to large semantic shifts. This undermines both confidence and interpretability, and suggests that correct scores do not necessarily reflect robust intermediate understanding.

Taken together, these findings reinforce the idea that Banff scores are not absolute ground truths but approximations grounded in expert consensus and subjective context. While AI pipelines offer scalable and reproducible outputs, fully replacing expert-level reasoning—especially for semi-quantitative lesion grading—remains an open challenge.

\section{Discussion}
The observed underestimation and misclassification patterns across $g$, $ptc$, and $v$ likely arise from multiple compounding factors rather than a single model deficiency. First, the pipeline’s dependency on upstream structural segmentation and inflammatory cell detection introduces vulnerability to error propagation. As shown in Figures~\ref{fig:structural-omission} and~\ref{fig:structural-hallucination}, missed anatomical regions or hallucinated structures directly alter the lesion context in which Banff scores are computed. Such instability is further amplified when the intermediate outputs—particularly cell detections—exhibit variability across runs or are sensitive to minor changes in preprocessing.

Second, annotation variability between pathologists contributes an intrinsic source of noise. Banff scoring involves semi-quantitative thresholds (e.g., a minimum number of inflammatory cells in a specific compartment), but in practice, morphological interpretation and compartment boundaries are often subjective. This means that even with perfect model replication of one annotator’s criteria, disagreement with another expert remains possible. The absence of large-scale, consensus-annotated datasets exacerbates this issue and limits the ability to learn a universally accepted scoring standard.

Third, the threshold-based nature of Banff definitions makes the grading process inherently sensitive to small perturbations in detection counts. As illustrated in Figure~\ref{fig:good-pred-but-uncertain}, the final score may be correct while the underlying detections include both false positives and false negatives. Near decision boundaries, such fluctuations can flip the assigned grade, undermining robustness and interpretability.

Finally, class imbalance—particularly the scarcity of high-grade lesions in training and evaluation—restricts the model’s exposure to rare but clinically significant patterns. Without sufficient examples, both structural and cellular detectors may fail to generalize to severe cases, reinforcing the tendency toward lower-grade predictions.

Overall, these findings indicate that current AI pipelines face fundamental challenges in Banff lesion scoring that extend beyond algorithmic accuracy. Addressing them will require reducing upstream instability, improving inter-observer consensus in annotations, designing grading schemes resilient to detection noise, and curating balanced datasets that capture the full spectrum of lesion severity.

\section{Conclusion}

In this study, we evaluated the feasibility of approximating Banff lesion scores using existing AI tools through a modular, rule-based framework. By decomposing each Banff indicator into its cellular and structural components, we systematically mapped which lesion types can be supported by current segmentation and detection models. Our findings suggest that while indicators like glomerulitis ($g$), peritubular capillaritis ($ptc$), and intimal arteritis ($v$) can be partially estimated using available tools, score accuracy remains limited by structural segmentation errors and detection ambiguities.

Importantly, our results show that even when predicted scores match expert annotations, the underlying reasoning may be inconsistent due to false positive or false negative detections. This raises critical concerns regarding interpretability and trustworthiness in AI-assisted scoring, particularly for indicators based on thresholded cell counts or subtle anatomical features. Failure cases further illustrate how structural omissions and hallucinations directly propagate to final score discrepancies.

Rather than proposing a complete AI-driven Banff scoring pipeline, our work highlights the importance of modular analysis in identifying the capabilities and blind spots of existing models. This component-level perspective enables targeted development of future models, data annotations, and validation strategies. Bridging the gap between algorithmic outputs and expert-level pathology requires not only improved model performance, but also closer alignment with the semi-quantitative, context-aware reasoning that defines Banff criteria.

We hope this feasibility-oriented approach can serve as a blueprint for systematically advancing AI-assisted pathology tools toward reliable, interpretable, and clinically aligned transplant biopsy evaluation.

\acknowledgements
\begin{flushleft}
This research was supported by NIH R01DK135597 (Huo), DoD HT9425-23-1-0003 (HCY), and KPMP Glue Grant. This work was also supported by Vanderbilt Seed Success Grant, Vanderbilt Discovery Grant, and VISE Seed Grant. This project was supported by The Leona M. and Harry B. Helmsley Charitable Trust grant G-1903-03793 and G-2103-05128. This research was also supported by NIH grants R01EB033385, R01DK132338, REB017230, R01MH125931, and NSF 2040462. We extend gratitude to NVIDIA for their support by means of the NVIDIA hardware grant. This work was also supported by NSF NAIRR Pilot Award NAIRR240055.
\end{flushleft}
\bibliographystyle{spiebib} 
\bibliography{report.bib}

\end{document}